\documentclass[twoside,11pt]{article}

\usepackage[abbrvbib, preprint, nohyperref]{jmlr2e}

\usepackage[most]{tcolorbox}
\usepackage{multicol}
\usepackage{amsmath}
\usepackage{amssymb}
\usepackage{xcolor}
\usepackage{color}
\usepackage{amsfonts}
\usepackage{mathtools}
\usepackage[bb=boondox,bbscaled=.95]{mathalfa}
\usepackage{bm}
\usepackage{physics}
\usepackage{cancel}
\usepackage{thmtools, thm-restate}
\usepackage{wrapfig}
\usepackage{marvosym}
\usepackage{pgfplots}
\usepackage{tikz}

\usepackage[frozencache, cachedir=.]{minted}
\definecolor{codegreen}{rgb}{0,0.6,0}
\definecolor{codegray}{rgb}{0.5,0.5,0.5}
\definecolor{codepurple}{rgb}{0.58,0,0.82}
\definecolor{backcolour}{rgb}{0.95,0.95,0.92}

\definecolor{olive}{rgb}{0.6, 0.6, 0.2}
\definecolor{sand}{rgb}{0.8666666666666667, 0.8, 0.4666666666666667}
\definecolor{wine}{rgb}{0.5333333333333333, 0.13333333333333333, 0.3333333333333333}
\definecolor{deblue}{RGB}{11,132,147}
\definecolor{ocra}{RGB}{204, 119, 34}
\definecolor{colone}{RGB}{209,220,204}
\definecolor{coltwo}{RGB}{204,222,210}
\definecolor{colthree}{RGB}{207,233,232}
\definecolor{boxt}{RGB}{11,111,147}

\newtcolorbox{codebox}[3][]
{
  colframe = deblue,
  colback  = #2!4,
  coltitle = #2!20!black,  
  title    = {#3},
  #1,
}
\def\td{{\tt TorchDyn}~}
\def\nnMod{{\tt torch.nn.Modules}~}

\newcommand{\fcircle}[2][red,fill=red]{\tikz[baseline=-0.5ex]\draw[#1,radius=#2] (0,0.03) circle ;}
\newtcolorbox{CatchyBox}[2][]{
    lower separated=false,
    colback=colthree,
    colframe=white, fonttitle=\bfseries,
    colbacktitle=coltwo,
    coltitle=black,
    enhanced,
    attach boxed title to top left={xshift=.02\linewidth,yshift=-4mm},
    title=#2,#1}

\usepgfplotslibrary{groupplots}
\usepgfplotslibrary{patchplots}
\pgfplotsset{compat=1.14} 

\definecolor{flow}{rgb}{0.5686274509803921, 0.5764705882352941, 0.7294117647058823}

\usetikzlibrary{matrix,positioning}

\definecolor{redi}{RGB}{255,38,0}
\definecolor{redii}{RGB}{200,50,30}
\definecolor{yellowi}{RGB}{255,251,0}
\definecolor{bluei}{RGB}{0,150,255}
\definecolor{blueii}{RGB}{135,247,210}
\definecolor{blueiii}{RGB}{91,205,250}
\definecolor{blueiv}{RGB}{115,244,253}
\definecolor{bluev}{RGB}{1,58,215}
\definecolor{orangei}{RGB}{240,143,50}
\definecolor{yellowii}{RGB}{222,247,100}
\definecolor{greeni}{RGB}{166,247,166}

\tikzset{ 
table/.style={
  matrix of nodes,
  row sep=-\pgflinewidth,
  column sep=-\pgflinewidth,
  nodes={rectangle,draw=black,text width=1.25ex,align=center},
  text depth=0.25ex,
  text height=1ex,
  nodes in empty cells
  },
texto/.style={font=\footnotesize\sffamily},
title/.style={font=\small\sffamily}
}

\newcommand\CellText[2]{%
  \node[texto,left=of mat#1,anchor=east]
  at (mat#1.west)
  {#2};
}

\newcommand\SlText[2]{%
  \node[texto,left=of mat#1,anchor=west,rotate=75]
  at ([xshift=3ex]mat#1.north)
  {#2};
}

\newcommand\RowTitle[2]{%
\node[title,left=6.3cm of mat#1,anchor=west]
  at (mat#1.north west)
  {#2};
}

\usepackage[colorlinks=true,linkcolor=blue,citecolor=blue]{hyperref}

\jmlrheading{1}{2020}{}{}{}{poli2020b}{Michael Poli et al.}

\ShortHeadings{TorchDyn: A PyTorch Library for Neural Differential Equations}{Poli, Massaroli, Yamashita, Asama and Park}
\firstpageno{1}

\begin{document}

\title{TorchDyn: A Neural Differential Equations Library}

\author{\name Michael Poli$^{1,\ddagger,\star}$ \email poli\_m@kaist.ac.kr\\
       \name Stefano Massaroli$^{2,\ddagger, \star}$ \email massaroli@robot.t.u-tokyo.ac.jp\\
       \name Atsushi Yamashita$^{2}$ \email yamashita@robot.t.u-tokyo.ac.jp\\
       \name Hajime Asama$^{2}$ \email asama@robot.t.u-tokyo.ac.jp\\
       \name Jinkyoo Park$^{1}$ \email jinkyoo.park@kaist.ac.kr
       \AND
       \addr $^1$KAIST, Daejeon, South Korea\\
       \addr $^2$University of Tokyo, Tokyo, Japan\\
       \addr $^\ddagger$DiffEqML\\
       \addr $^\star$Equal Contribution Authors
       }
\maketitle
\begin{abstract}
\textit{Continuous--depth} learning has recently emerged as a novel perspective on deep learning, improving performance in tasks related to dynamical systems and density estimation. Core to these approaches is the \textit{neural differential equation}, whose forward passes are the solutions of an initial value problem parametrized by a neural network. Unlocking the full potential of continuous--depth models requires a different set of software tools, due to peculiar differences compared to standard discrete neural networks, e.g inference must be carried out via numerical solvers. We introduce {\tt TorchDyn}, a PyTorch library dedicated to continuous--depth learning, designed to elevate neural differential equations to be as accessible as regular plug--and--play deep learning primitives. This objective is achieved by identifying and subdividing different variants into common essential components, which can be combined and freely repurposed to obtain complex compositional architectures. {\tt TorchDyn} further offers step--by--step tutorials and benchmarks designed to guide researchers and contributors.
\end{abstract}
\section{Introduction}
With foundational work now decades old \citep{cohen1983absolute,hopfield1984neurons,lecun1988theoretical,zhang2014comprehensive}, the blend of differential equations, deep learning and dynamical systems has been reignited by recent works on a novel computational primitive: \textit{neural differential equations} \citep{chen2018neural}. Often referred to as \textit{continuous--depth learning}, this new paradigm has shown promise across a plethora of different machine learning tasks, such as density estimation \citep{chen2018neural,grathwohl2018ffjord}, forecasting \citep{rubanova2019latent,poli2019graph,portwood2019turbulence}, time series classification \citep{kidger2020neural}, image segmentation \citep{pinckaers2019neural}.\\
Continuous--depth models rely on additional \textit{machinery} and supporting modules not present in standard deep learning libraries. Indeed, inspecting modern software implementations of state--of--the--art variants in the literature of neural differential equations reveals a prohibitive amount of boilerplate code, with convoluted inheritance chains\footnote{An example is the original FFJORD \citep{grathwohl2018ffjord}, where long inheritance chains are present due to intertwining of model details and boilerplate classes. {\tt TorchDyn} preserves modularity by disentangling these factors, significantly simplifying extensions or modifications to the original model.}. As a result, advances in the field are hindered to be both slower and less reproducible, leading to longer onboarding times for researchers and practitioners interested in deploying neural differential equations in and outside academia. {\tt TorchDyn} aims at filling these gaps, by providing a fully--featured library for continuous--depth models. Through {\tt TorchDyn} neural differential equations and derivative models, e.g \citep{greydanus2019hamiltonian,toth2019hamiltonian,massaroli2020dissecting,lutter2019deep,cranmer2020lagrangian,massaroli2020stable, li2020scalable}, including \textit{yet--to--be--published} combinations, can effortlessly be obtained by \textit{ad hoc} primitives in combination with the rich {\tt PyTorch} \citep{paszke2019pytorch} ecosystem. 
\begin{wrapfigure}[10]{r}{0.425\textwidth}\label{fig:core_ele}
  \centering
  \vspace*{4mm}
  \includegraphics[width=0.4\textwidth]{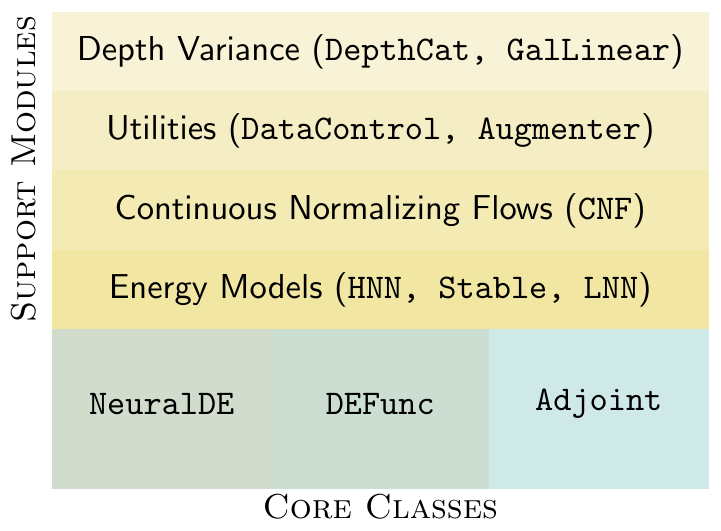}
  \vspace*{-5mm}
  \caption{\footnotesize{Core elements of {\tt TorchDyn}}.}
\end{wrapfigure}
\section{The {\tt TorchDyn} Library}
\paragraph{Design philosophy}
The main objective of {\tt TorchDyn} is to offer a complete, intuitive access--point to the continuous--depth framework, which can be interfaced with {\tt PyTorch} to obtain architectures going beyond modern literature on the topic. We follow core design ideals driving the success of modern deep learning frameworks such as {\tt PyTorch}; namely, modular, object--oriented, and with a focus on GPUs and batched operations. 
\paragraph{Software dependencies}
{\tt TorchDyn} is embedded in {\tt Python} and is built to be compatible with {\tt PyTorch}, relying on a select number of supporting libraries. We utilize {\tt torchdiffeq} \citep{chen2018neural} and {\tt torchsde} \citep{li2020scalable}, along with recent learning--based approaches \citep{poli2020hypersolvers} as a source of numerical methods for differential equations, augmented by {\tt PyTorch-Lightning} \citep{falcon2019pytorch} for logging and training loops. We note that {\tt TorchDyn} is fully-functional without {\tt PyTorch-Lightning}, included to improve quality of life and reduce boilerplate for practitioners who do not wish to work at a higher level of abstraction that native {\tt PyTorch}.
\section{The Elements of {\tt TorchDyn}}
We detail the core elements of the library highlighted in Figure \ref{fig:core_ele} via specific examples.
\paragraph{NeuralDE and DEFunc}
At the core of {\tt TorchDyn} lie the {\tt NeuralDE} and {\tt DEFunc} classes. {\tt NeuralDE}s represent the primary model class which can interacted with in usual {\tt PyTorch} fashion. Internally, {\tt DEFunc} perform auxiliary operations required to preserve compatibility across {\tt NeuralDE} variants, such as higher--order dynamics or handling additional dimensions for integral cost functions. 
\paragraph{Utilities and depth--variance}
Neural differential equations can be defined to have parameters either fixed or varying in depth. This is achieved in a different way than regular discrete neural networks, which store a separate parameter tensor for each layer and index them with the appropriate layer index. This methodology would be impossible for continuous models as the depth--variable indexing the parameters takes real values, and as such leads to an infinite number of layers and thus parameters. There exist several strategies to sidestep this issue; approximating depth--variance by concatenating to the state, usually achieved by defining modified {\tt torch.nn.Module}s, or utilizing spectral or depth discretizations \citep{massaroli2020dissecting}. {\tt TorchDyn} includes dedicated layers for depth--variance: {\tt DepthCat} and {\tt GalLinear, GalConv2d}, which eliminate the need of defining additional modules to achieve the desired effect.
\paragraph{Energy models} There exists a whole line of work of physics--inspired Neural ODE variants such as Hamiltonian Neural Networks \citep{greydanus2019hamiltonian}, Lagrangian Neural Networks \citep{lutter2019deep, cranmer2020lagrangian} or general energy--based models \citep{massaroli2020stable}. \td  fully supports these models by dedicated wrappers for any {\nnMod} parametrizing the Energy/Hamiltonian/Lagrangian function. These wrappers are further preserved to be compatible with other modules by {\tt DEFunc}s, allowing several combinations as shown in Figure \ref{fig:all_combos}.
\paragraph{Continuous normalizing flows}
An additional fundamental member of the continuous--depth framework, \textit{continuous normalizing flows} (CNFs) \citep{chen2018neural,grathwohl2018ffjord} are treated as first--class primitives. Notably, this allows for an out--of--the--box definition of Hamiltonian CNFs \citep{toth2019hamiltonian} and other unpublished variants.
\paragraph{Adjoint and integral losses}
\td implements a complete sensitivity toolset for Neural ODEs. This includes standard back--propagation through the ODE solver ($\mathcal{O}(\tilde S)$ memory efficiency\footnote{$\tilde{S}$ represents the number of integration step of the numerical solver.}) and \textit{adjoint}--based gradients ($\mathcal{O}(1)$ memory efficiency). Compared to {\tt torchdiffeq}'s adjoint method, \td offers the capability to compute gradients of \textit{integral}\footnote{Integral loss functions are defined on the whole depth domain, e.g. $\ell := \displaystyle\int_{\mathcal{S}} l(\theta, {\bf z}(\tau), \tau)\dd\tau$. See \citep{massaroli2020dissecting} for further details on the derivation of the adjoint method for this type of losses.} loss functions which are gaining increasing popularity in the Neural ODEs research community \citep{massaroli2020stable, finlay2020train}.
\subsection{Tutorials and documentation}
The continuous--depth framework requires a set of background theoretical developments less familiar to deep learning specialists. {\tt TorchDyn} offers a series of step--by--step, tutorial notebooks guiding new users and contributors. Some examples include a complete {\tt cookbook} for neural ordinary differential equation variants, and tutorials for Hamiltonian Neural Networks \citep{greydanus2019hamiltonian}, FFJORD \citep{grathwohl2018ffjord}, Neural Graph Ordinary Differential Equations \citep{poli2019graph} and more.
\begin{figure}
\makeatletter
\makeatother
\begin{minipage}[h]{0.42\linewidth}
\begin{minted}
[
framesep=1mm,
baselinestretch=1,
fontsize=\scriptsize,
linenos
]
{python}
from torchdyn import *
import torch.nn as nn
# vector field parametrized by a NN
f = nn.Sequential(DepthCat(1),
                  nn.Linear(3, 64),
                  nn.Tanh(), 
                  nn.Linear(64, 2))

model = NeuralODE(f, 
                  sensitivity='adjoint', 
                  solver='dopri5')

x = torch.rand(128, 2)
print(model(x).shape)
>> torch.Size([128, 2])

x = torch.rand(128, 2)
s_span = torch.linspace(0, 2, 50)
print(model.trajectory(x, s_span).shape)
>> torch.Size([50, 128, 2])

\end{minted}
\vfill
\end{minipage}
\hfill\vline\hfill
\begin{minipage}[h]{.53\linewidth}\small
\begin{minted}
[
framesep=1mm,
baselinestretch=1,
fontsize=\scriptsize,
linenos
]
{python}
print(model)
            
Neural DE:
	- order: 1        
	- solver: dopri5
	- integration interval: 0.0 to 1.0        
	- num_checkpoints: 2        
	- tolerances: relative 0.0001 absolute 0.0001        
	- num_parameters: 386        
	- NFE: 40.0
        
Integral loss: None
        
DEFunc:
 DEFunc(
  (m): Sequential(
    (0): DepthCat()
    (1): Linear(in_features=3, out_features=64, bias=True)
    (2): Tanh()
    (3): Linear(in_features=64, out_features=2, bias=True)
  )
)
\end{minted}
\vfill
\end{minipage}
\vspace*{-4mm}
\caption{\footnotesize Examples of the flexible {\tt NeuralDE} API, compatible with {\tt PyTorch}. The Neural ODE can instance can be passed tensor data directly, implicitly utilizing the {\tt .forward}. Alternatively, the ODE can be evaluated at a predetermined set of depth points via a {\tt .trajectory} method, resulting in an output of dimensions \textit{length, batch size, data dimension}.}
\vspace*{-10mm}
\end{figure}

\section{Related Software}
 To the best knowledge of the authors, {\tt TorchDyn} represents the first {\tt Python}--based library entirely dedicated to continuous--depth models and utilities. {\tt TorchDyn} builds upon {\tt torchdiffeq} \citep{chen2018neural} and {\tt torchsde} \citep{li2020scalable} as a source of numerical solvers for batched ODEs and SDEs on GPUs. Outside of {\tt Python}, the {\tt SciML} ecosystem provides a {\tt Julia}--based alternative with a primary focus on scientific machine learning and smaller data regimes. {\tt DiffEqFlux} \citep{rackauckas2019diffeqflux} provides some of the model classes present in {\tt TorchDyn}, such as ANODEs \citep{dupont2019augmented} and FFJORD \citep{grathwohl2018ffjord}, though with less emphasis on modularity and composability. {\tt DiffEqFlux} relies on {\tt Flux} for its deep learning primitives, which is notably more limited and less established in its current form than Python--based frameworks such as {\tt PyTorch} or {\tt TensorFlow}. The advantage for {\tt DiffEqFlux} in specialized scientific application is due to the extensive numerical method suite offered by {\tt SciML} \citep{rackauckas2017differentialequations}.
\begin{figure}[h]\label{fig:all_combos}
    \centering
    \vspace*{-7mm}
    \scalebox{0.75}{\begin{tikzpicture}[node distance =0pt and 0.5cm]

\matrix[table, nodes in empty cells] (mat11) 
{
  |[fill=greeni]| & |[fill=greeni]|\\
  |[fill=greeni]| & |[fill=greeni]|\\
  |[fill=greeni]| & |[fill=greeni]|\\
  |[fill=greeni]| & |[fill=greeni]|\\
  |[fill=yellowi]| \tiny{$\mathcal H$}& |[fill=greeni]|\\
  |[fill=yellowi]|\tiny{$\mathcal L$}& |[fill=greeni]|\\
  |[fill=bluei]|  & |[fill=greeni]|\\
  |[fill=greeni]| & |[fill=greeni]|\\
  |[fill=greeni]| & |[fill=yellowi]| \tiny ad\\
  |[fill=greeni]| & |[fill=yellowi]| \tiny h\\
  |[fill=greeni]| & |[fill=yellowi]|\tiny h\\
};
\matrix[table,right=of mat11] (mat12)
{
  |[fill=bluei]| & |[fill=bluei]| & |[fill=greeni]|\\
  |[fill=greeni]|& |[fill=bluei]| & |[fill=greeni]|\\
  |[fill=greeni]|& |[fill=bluei]| & |[fill=bluei]|\\
  |[fill=greeni]|& |[fill=bluei]| & |[fill=greeni]|\\
  |[fill=greeni]|& |[fill=greeni]|& |[fill=greeni]|\\
  |[fill=greeni]|& |[fill=greeni]|& |[fill=greeni]|\\
  |[fill=bluei]| & |[fill=greeni]|& |[fill=bluei]|\\
  |[fill=greeni]|& |[fill=bluei]| & |[fill=bluei]|\\
  |[fill=greeni]|& |[fill=bluei]| & |[fill=greeni]|\\
  |[fill=greeni]|& |[fill=bluei]| & |[fill=greeni]|\\
  |[fill=greeni]|& |[fill=bluei]| & |[fill=greeni]|\\
};
\matrix[table,right=of mat12] (mat13) 
{
  |[fill=bluei]| & |[fill=greeni]|\\
  |[fill=bluei]| & |[fill=greeni]|\\
  |[fill=bluei]| & |[fill=bluei]|\\
  |[fill=bluei]| & |[fill=greeni]|\\
  |[fill=greeni]|& |[fill=greeni]|\\
  |[fill=greeni]|& |[fill=greeni]|\\
  |[fill=bluei]| & |[fill=bluei]|\\
  |[fill=bluei]| & |[fill=greeni]|\\
  |[fill=bluei]| & |[fill=bluei]|\\
  |[fill=bluei]| & |[fill=bluei]|\\
  |[fill=bluei]| & |[fill=bluei]|\\
};

\SlText{11-1-1}{Energy}
\SlText{11-1-2}{CNF}

\SlText{12-1-1}{Augmentation}
\SlText{12-1-2}{Depth--Variance}
\SlText{12-1-3}{Data--Control}

\SlText{13-1-1}{Adjoint}
\SlText{13-1-2}{Integral Loss}

\RowTitle{11}{Model};
\CellText{11-1-1}{ANODE \citep{dupont2019augmented}};
\CellText{11-2-1}{Higher-Order \citep{massaroli2020dissecting}};
\CellText{11-3-1}{Gal\"erkin Neural ODEs \citep{massaroli2020dissecting}};
\CellText{11-4-1}{Stacked Neural ODEs \citep{massaroli2020dissecting}};
\CellText{11-5-1}{Hamiltonian \citep{greydanus2019hamiltonian}};
\CellText{11-6-1}{Lagrangian \citep{lutter2019deep, cranmer2020lagrangian}};
\CellText{11-7-1}{Stable Neural Flows \citep{massaroli2020stable}};
\CellText{11-8-1}{Graph Neural ODEs \citep{poli2019graph}};
\CellText{11-9-1}{CNF \citep{chen2018neural}};
\CellText{11-10-1}{FFJORD \citep{grathwohl2018ffjord}};
\CellText{11-11-1}{RNODE \citep{finlay2020train}};

\end{tikzpicture}}
    \vspace*{-5mm}
    \caption{\footnotesize Support in {\tt torchdyn} of different SOTA models. \fcircle[black, fill=greeni]{4pt}: fully supported as the original paper. \fcircle[black, fill=bluei]{4pt}: supported in {\tt torchdyn} but not present in the original paper. \fcircle[black, fill=yellowi]{4pt}\hspace{-2.5mm}{\tiny$\mathcal{H}$} : Hamiltonian--type energy functions. \fcircle[black, fill=yellowi]{4pt}\hspace{-2.5mm}{\tiny$\mathcal{L}$} : Lagrangian--type energy functions. \fcircle[black, fill=yellowi]{4pt}\hspace{-2.5mm}{\tiny{\tt ad}} : autograd trace. \fcircle[black, fill=yellowi]{4pt}\hspace{-2mm}{\vspace*{-2mm}\tiny{\tt h}} : Hutchinson estimator.}
    \label{fig:my_label}
    \vspace*{-10mm}
\end{figure}
\section{Conclusions}
We introduce {\tt TorchDyn}, the first fully--featured library for continuous--depth models compatible with {\tt PyTorch}. {\tt TorchDyn} is designed to provide an intuitive, modular and powerful interface for neural differential equations and derivative models. The library further provides a complete set of tutorials to guide new practitioners and researchers. 
\bibliography{biblio.bib}

\begin{thebibliography}{25}
\providecommand{\natexlab}[1]{#1}
\providecommand{\url}[1]{\texttt{#1}}
\expandafter\ifx\csname urlstyle\endcsname\relax
  \providecommand{\doi}[1]{doi: #1}\else
  \providecommand{\doi}{doi: \begingroup \urlstyle{rm}\Url}\fi

\bibitem[Chen et~al.(2018)Chen, Rubanova, Bettencourt, and
  Duvenaud]{chen2018neural}
R.~T. Chen, Y.~Rubanova, J.~Bettencourt, and D.~K. Duvenaud.
\newblock Neural ordinary differential equations.
\newblock In \emph{Advances in neural information processing systems}, pages
  6571--6583, 2018.

\bibitem[Cohen and Grossberg(1983)]{cohen1983absolute}
M.~A. Cohen and S.~Grossberg.
\newblock Absolute stability of global pattern formation and parallel memory
  storage by competitive neural networks.
\newblock \emph{IEEE transactions on systems, man, and cybernetics}, \penalty0
  (5):\penalty0 815--826, 1983.

\bibitem[Cranmer et~al.(2020)Cranmer, Greydanus, Hoyer, Battaglia, Spergel, and
  Ho]{cranmer2020lagrangian}
M.~Cranmer, S.~Greydanus, S.~Hoyer, P.~Battaglia, D.~Spergel, and S.~Ho.
\newblock Lagrangian neural networks.
\newblock \emph{arXiv preprint arXiv:2003.04630}, 2020.

\bibitem[Dupont et~al.(2019)Dupont, Doucet, and Teh]{dupont2019augmented}
E.~Dupont, A.~Doucet, and Y.~W. Teh.
\newblock Augmented neural odes.
\newblock In \emph{Advances in Neural Information Processing Systems}, pages
  3134--3144, 2019.

\bibitem[Falcon et~al.(2019)]{falcon2019pytorch}
W.~Falcon et~al.
\newblock Pytorch lightning.
\newblock \emph{GitHub. Note: https://github.
  com/williamFalcon/pytorch-lightning Cited by}, 3, 2019.

\bibitem[Finlay et~al.(2020)Finlay, Jacobsen, Nurbekyan, and
  Oberman]{finlay2020train}
C.~Finlay, J.-H. Jacobsen, L.~Nurbekyan, and A.~M. Oberman.
\newblock How to train your neural ode.
\newblock \emph{arXiv preprint arXiv:2002.02798}, 2020.

\bibitem[Grathwohl et~al.(2018)Grathwohl, Chen, Bettencourt, Sutskever, and
  Duvenaud]{grathwohl2018ffjord}
W.~Grathwohl, R.~T. Chen, J.~Bettencourt, I.~Sutskever, and D.~Duvenaud.
\newblock Ffjord: Free-form continuous dynamics for scalable reversible
  generative models.
\newblock \emph{arXiv preprint arXiv:1810.01367}, 2018.

\bibitem[Greydanus et~al.(2019)Greydanus, Dzamba, and
  Yosinski]{greydanus2019hamiltonian}
S.~Greydanus, M.~Dzamba, and J.~Yosinski.
\newblock Hamiltonian neural networks.
\newblock In \emph{Advances in Neural Information Processing Systems}, pages
  15379--15389, 2019.

\bibitem[Hopfield(1984)]{hopfield1984neurons}
J.~J. Hopfield.
\newblock Neurons with graded response have collective computational properties
  like those of two-state neurons.
\newblock \emph{Proceedings of the national academy of sciences}, 81\penalty0
  (10):\penalty0 3088--3092, 1984.

\bibitem[Kidger et~al.(2020)Kidger, Morrill, Foster, and
  Lyons]{kidger2020neural}
P.~Kidger, J.~Morrill, J.~Foster, and T.~Lyons.
\newblock Neural controlled differential equations for irregular time series.
\newblock \emph{arXiv preprint arXiv:2005.08926}, 2020.

\bibitem[LeCun et~al.(1988)LeCun, Touresky, Hinton, and
  Sejnowski]{lecun1988theoretical}
Y.~LeCun, D.~Touresky, G.~Hinton, and T.~Sejnowski.
\newblock A theoretical framework for back-propagation.
\newblock In \emph{Proceedings of the 1988 connectionist models summer school},
  volume~1, pages 21--28. CMU, Pittsburgh, Pa: Morgan Kaufmann, 1988.

\bibitem[Li et~al.(2020)Li, Wong, Chen, and Duvenaud]{li2020scalable}
X.~Li, T.-K.~L. Wong, R.~T. Chen, and D.~Duvenaud.
\newblock Scalable gradients for stochastic differential equations.
\newblock \emph{arXiv preprint arXiv:2001.01328}, 2020.

\bibitem[Lutter et~al.(2019)Lutter, Ritter, and Peters]{lutter2019deep}
M.~Lutter, C.~Ritter, and J.~Peters.
\newblock Deep lagrangian networks: Using physics as model prior for deep
  learning.
\newblock \emph{arXiv preprint arXiv:1907.04490}, 2019.

\bibitem[Massaroli et~al.(2020{\natexlab{a}})Massaroli, Poli, Bin, Park,
  Yamashita, and Asama]{massaroli2020stable}
S.~Massaroli, M.~Poli, M.~Bin, J.~Park, A.~Yamashita, and H.~Asama.
\newblock Stable neural flows.
\newblock \emph{arXiv preprint arXiv:2003.08063}, 2020{\natexlab{a}}.

\bibitem[Massaroli et~al.(2020{\natexlab{b}})Massaroli, Poli, Park, Yamashita,
  and Asama]{massaroli2020dissecting}
S.~Massaroli, M.~Poli, J.~Park, A.~Yamashita, and H.~Asama.
\newblock Dissecting neural odes.
\newblock \emph{arXiv preprint arXiv:2002.08071}, 2020{\natexlab{b}}.

\bibitem[Paszke et~al.(2019)Paszke, Gross, Massa, Lerer, Bradbury, Chanan,
  Killeen, Lin, Gimelshein, Antiga, et~al.]{paszke2019pytorch}
A.~Paszke, S.~Gross, F.~Massa, A.~Lerer, J.~Bradbury, G.~Chanan, T.~Killeen,
  Z.~Lin, N.~Gimelshein, L.~Antiga, et~al.
\newblock Pytorch: An imperative style, high-performance deep learning library.
\newblock In \emph{Advances in neural information processing systems}, pages
  8026--8037, 2019.

\bibitem[Pinckaers and Litjens(2019)]{pinckaers2019neural}
H.~Pinckaers and G.~Litjens.
\newblock Neural ordinary differential equations for semantic segmentation of
  individual colon glands.
\newblock \emph{arXiv preprint arXiv:1910.10470}, 2019.

\bibitem[Poli et~al.(2019)Poli, Massaroli, Park, Yamashita, Asama, and
  Park]{poli2019graph}
M.~Poli, S.~Massaroli, J.~Park, A.~Yamashita, H.~Asama, and J.~Park.
\newblock Graph neural ordinary differential equations.
\newblock \emph{arXiv preprint arXiv:1911.07532}, 2019.

\bibitem[Poli et~al.(2020)Poli, Massaroli, Yamashita, Asama, and
  Park]{poli2020hypersolvers}
M.~Poli, S.~Massaroli, A.~Yamashita, H.~Asama, and J.~Park.
\newblock Hypersolvers: Toward fast continuous-depth models.
\newblock \emph{arXiv preprint arXiv:2007.09601}, 2020.

\bibitem[Portwood et~al.(2019)Portwood, Mitra, Ribeiro, Nguyen, Nadiga, Saenz,
  Chertkov, Garg, Anandkumar, Dengel, et~al.]{portwood2019turbulence}
G.~D. Portwood, P.~P. Mitra, M.~D. Ribeiro, T.~M. Nguyen, B.~T. Nadiga, J.~A.
  Saenz, M.~Chertkov, A.~Garg, A.~Anandkumar, A.~Dengel, et~al.
\newblock Turbulence forecasting via neural ode.
\newblock \emph{arXiv preprint arXiv:1911.05180}, 2019.

\bibitem[Rackauckas and Nie(2017)]{rackauckas2017differentialequations}
C.~Rackauckas and Q.~Nie.
\newblock Differentialequations. jl--a performant and feature-rich ecosystem
  for solving differential equations in julia.
\newblock \emph{Journal of Open Research Software}, 5\penalty0 (1), 2017.

\bibitem[Rackauckas et~al.(2019)Rackauckas, Innes, Ma, Bettencourt, White, and
  Dixit]{rackauckas2019diffeqflux}
C.~Rackauckas, M.~Innes, Y.~Ma, J.~Bettencourt, L.~White, and V.~Dixit.
\newblock Diffeqflux. jl-a julia library for neural differential equations.
\newblock \emph{arXiv preprint arXiv:1902.02376}, 2019.

\bibitem[Rubanova et~al.(2019)Rubanova, Chen, and Duvenaud]{rubanova2019latent}
Y.~Rubanova, T.~Q. Chen, and D.~K. Duvenaud.
\newblock Latent ordinary differential equations for irregularly-sampled time
  series.
\newblock In \emph{Advances in Neural Information Processing Systems}, pages
  5321--5331, 2019.

\bibitem[Toth et~al.(2019)Toth, Rezende, Jaegle, Racani{\`e}re, Botev, and
  Higgins]{toth2019hamiltonian}
P.~Toth, D.~J. Rezende, A.~Jaegle, S.~Racani{\`e}re, A.~Botev, and I.~Higgins.
\newblock Hamiltonian generative networks.
\newblock \emph{arXiv preprint arXiv:1909.13789}, 2019.

\bibitem[Zhang et~al.(2014)Zhang, Wang, and Liu]{zhang2014comprehensive}
H.~Zhang, Z.~Wang, and D.~Liu.
\newblock A comprehensive review of stability analysis of continuous-time
  recurrent neural networks.
\newblock \emph{IEEE Transactions on Neural Networks and Learning Systems},
  25\penalty0 (7):\penalty0 1229--1262, 2014.

\end{thebibliography}
\end{document}